%% file: aaai2026.tex
\documentclass[letterpaper]{article} 
\usepackage{aaai2026}  
\usepackage{times}  
\usepackage{helvet}  
\usepackage{courier}  
\usepackage[hyphens]{url}  
\usepackage{graphicx} 
\usepackage{natbib}  
\usepackage{caption} 
\usepackage{algorithm}
\usepackage{algorithmic}
\usepackage{newfloat}
\usepackage{listings}
\DeclareCaptionStyle{ruled}{labelfont=normalfont,labelsep=colon,strut=off} 
\floatstyle{ruled}
\newfloat{listing}{tb}{lst}{}
\floatname{listing}{Listing}
\usepackage{times}
\usepackage{helvet}
\usepackage{courier}
\usepackage{enumitem}
\usepackage{graphicx}
\usepackage{amsfonts}
\usepackage{amsmath} 
\usepackage{booktabs} 
\usepackage{multirow} 
\usepackage{xspace}
\usepackage{color}
\usepackage{colortbl}
\definecolor{truecolor}{RGB}{221,160,221}
\definecolor{wrongcolor}{RGB}{122,129,255}
\definecolor{uncolor}{RGB}{255,147,0}

\title{Multimodal Mixture-of-Experts with Retrieval Augmentation for Protein Active Site Identification}

\usepackage{bibentry}

\begin{document}
%


\author{
Jiayang Wu\textsuperscript{\rm 1}\thanks{Equal contribution.},
Jiale Zhou\textsuperscript{\rm 1}\footnotemark[1],
Rubo Wang\textsuperscript{\rm 2}\thanks{Corresponding author.},
Xingyi Zhang\textsuperscript{\rm 3},
Xun Lin\textsuperscript{\rm 1},\\
Tianxu Lv\textsuperscript{\rm 4},
Leong Hou U\textsuperscript{\rm 5},
Yefeng Zheng\textsuperscript{\rm 1}\footnotemark[2]
}
\affiliations{
\textsuperscript{\rm 1}Westlake University
\textsuperscript{\rm 2}Chinese Academy of Sciences\\
\textsuperscript{\rm 3}Mohamed bin Zayed University of Artificial Intelligence\\
\textsuperscript{\rm 4}Jiangnan University
\textsuperscript{\rm 5}University of Macau
}


\maketitle

\input{section/0-abstract}

\input{section/1-intro}

\input{section/2-related_work}

\input{section/3-method}

\input{section/4-experiments}

\input{section/5-conclusion}

\section*{Acknowledgements}
This work was supported by Zhejiang Leading Innovative and Entrepreneur Team Introduction Program (2024R01007).

\bibliography{aaai2026}

\end{document}

%% file: section/0-abstract.tex
\begin{abstract}

Accurate identification of protein active sites at the residue level is crucial for understanding protein function and advancing drug discovery. However, current methods face two critical challenges: vulnerability in single-instance prediction due to sparse training data, and inadequate modality reliability estimation that leads to performance degradation when unreliable modalities dominate fusion processes. To address these challenges, we introduce \textbf{M}ultimodal Mixture-of-\textbf{E}xperts with \textbf{R}etrieval \textbf{A}ugmentation (MERA), the first retrieval-augmented framework for protein active site identification. MERA employs hierarchical multi-expert retrieval that dynamically aggregates contextual information from chain, sequence, and active-site perspectives through residue-level mixture-of-experts gating. To prevent modality degradation, we propose a reliability-aware fusion strategy based on Dempster–Shafer evidence theory that quantifies modality trustworthiness through belief mass functions and learnable discounting coefficients, enabling principled multimodal integration. Extensive experiments on ProTAD-Gen and TS125 datasets demonstrate that MERA achieves state-of-the-art performance, with 90\% AUPRC on active site prediction and significant gains on peptide-binding site identification, validating the effectiveness of retrieval-augmented multi-expert modeling and reliability-guided fusion. Code is available at \url{https://github.com/csjywu1/MERA}.

\end{abstract}

%% file: section/1-intro.tex
\section{Introduction}

Accurate residue-level active site identification remains a critical bottleneck in mechanistic biology and drug discovery. This challenge stems fundamentally from the extreme label sparsity of catalytic/binding residues, which constitute less than 0.5\% of all protein positions, compounded by significant functional divergence across homologous families~\cite{petrova2006prediction}. Consequently, despite decades of research, these functionally critical residues continue to elude precise prediction. This fundamental scarcity and resulting prediction inaccuracy directly impede the effectiveness of virtual screening pipelines~\cite{gligorijevic2021structure}.

To overcome the dual bottlenecks of scarce annotations and extreme class imbalance, current research has largely pursued three complementary paradigms: (i) harnessing \textbf{sequence-only} pre-training to transfer massive unlabeled evolutionary knowledge, as demonstrated by ESM-1b \cite{rives2021biological} and ProtTrans \cite{elnaggar2021prottrans}; (ii) incorporating sparse but precise 3D cues via \textbf{structure-aware} encoders when experimental coordinates are available, such as MIF \cite{yang2023masked} and PST \cite{chen2024endowing}; and (iii) enhancing predictive capacity via \textbf{cross-modal fusion} that integrates auxiliary modalities like protein textual descriptions through sequence-text embedding alignment, including MMSite \cite{ouyang2024mmsite}, UniSite \cite{fan2025unisite}, and ProtST \cite{xu2023protst}. Despite substantial progress, two persistent challenges persist in achieving reliable active-site prediction.

\begin{figure}[!t]
    \centering
    \includegraphics[clip,width=0.95\linewidth]{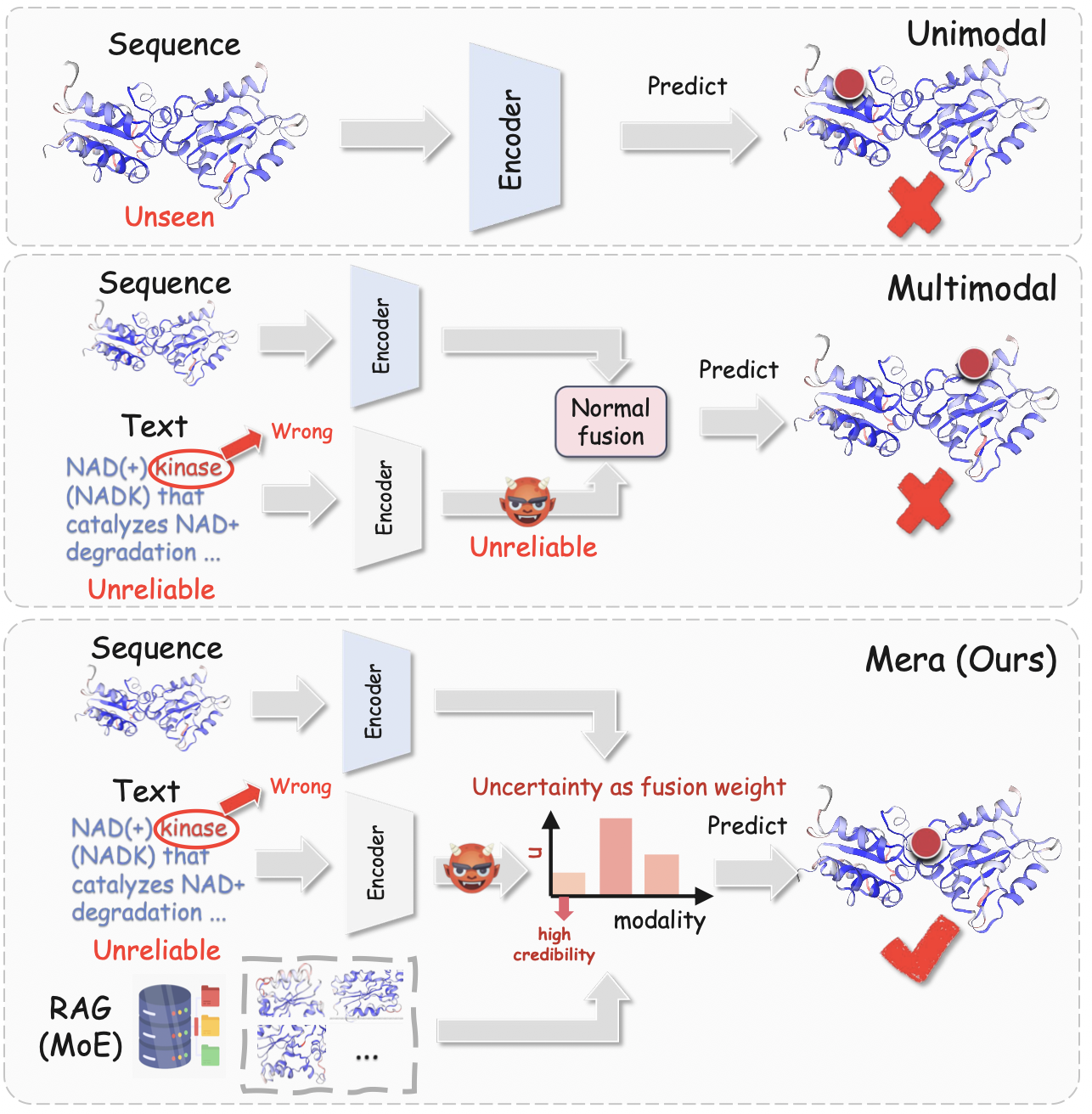}
    \vspace{-2mm}
    \caption{Identification of protein active sites with sequence, text, and retrieval-augmented mixture-of-experts.}
    \vspace{-8mm}
    \label{fig:task}
\end{figure}

\textbf{Challenge 1: Vulnerability of single-instance prediction.} Predictions relying solely on intrinsic sequence features are inherently fragile due to sparse training data. This unreliability is particularly pronounced for rare protein sequences. While retrieval-augmented generation (RAG)~\cite{guu2020retrieval} offers a promising solution by incorporating external contextual information, naive sequence-level homolog retrieval often introduces noise that overwhelms informative signals due to the sparsity of active site. Consequently, extracting and integrating key information from diverse perspectives becomes crucial to accommodate sequence length variability and functional diversity across homologs.

\textbf{Challenge 2: Inadequate modality reliability estimation.} Existing fusion approaches typically assess modality contributions through learned attention weights or MLP-based coefficients, which serve as poor proxies for true modality reliability \cite{han2022multimodal,zhang2024multimodal,yang2024test}. The fundamental limitation lies in conflating \textit{contribution magnitude} with \textit{epistemic reliability}: both cross-attention and MLP-based fusion methods optimize for signal strength rather than modality trustworthiness, making them unreliable indicators for modality confidence. When unreliable modalities dominate the fusion process, substantial performance degradation can occur. A principled approach should explicitly distinguish between modality contribution strength and modality trustworthiness by modeling epistemic reliability at the residue level~\cite{shen2025mf}.

To address these challenges, we introduce \textbf{M}ultimodal Mixture-of-\textbf{E}xperts with \textbf{R}etrieval \textbf{A}ugmentation (\textbf{MERA}), a novel framework for protein active site identification. Our approach incorporates two key innovations. Multi-expert RAG (MeRAG) addresses the vulnerability of single-instance prediction by extracting meaningful information from sparse retrieval sequences through three orthogonal retrieval experts (chain, sequence, and active-site). Each expert  captures distinct biological information, and their outputs are dynamically aggregated via a residue-level Mixture-of-Experts (MoE) gating mechanism. This design enables adaptive and robust feature learning even under sparse data conditions. 
(ii) Reliability-aware Multimodal Fusion (RMF) addresses inadequate modality reliability estimation by explicitly quantifying the trustworthiness of each modality. Inspired by Dempster–Shafer evidence theory \cite{amini2020deep,huang2025deep,deregnaucourt2025conflict}, we model the prediction of each modality as a belief mass function and apply learnable discounting coefficients to reflect modality reliability. This principled approach enables the model to appropriately attenuate less reliable modalities during the fusion process, offering more robust multi-modal integration when information quality varies across modalities~\cite{han2024fmfn}. Guided by these reliability scores, we perform residue-level fusion to achieve more trustworthy predictions.

Our main contributions are summarized as follows:
\begin{itemize}
    \item We introduce MERA, the first retrieval-augmented framework for protein active site identification that employs residue-level MoE to dynamically retrieve and fuse contextual information from sequence, residue, and active-site views, yielding fine-grained enhancements tailored to individual residues.
    \item We propose a reliability-aware fusion strategy based on Dempster–Shafer evidence theory that quantifies modality trustworthiness through belief mass functions and learnable discounting coefficients, enabling principled multimodal integration for more robust active-site predictions.
    \item Comprehensive experiments validate the effectiveness of MERA on protein active site identification benchmarks, with additional protein-peptide binding site recognition experiments highlighting strong generalizability to more complex biological scenarios.
\end{itemize}

%% file: section/2-related_work.tex
\section{Related Work}

\subsection{Protein active site identification}
Protein active sites are spatially clustered residues within proteins that are directly responsible for molecular recognition \cite{nussinov2025pioneer}, substrate binding \cite{ji2024substrate} and catalyzing enzymatic reactions \cite{reisenbauer2024catalyzing}. Accurate identification of these functionally critical regions is fundamental to mechanistic biology and therapeutic design, as they directly determine protein structural and functional roles. Existing methods for protein active site identification can be broadly categorized into single-modality and multi-modal approaches.

\paragraph{Single-modality models} form the foundation of most current work and encompass both sequence-based and structure-based methods. Sequence-based models such as ESM-1b \cite{rives2021biological}, ESM-1v \cite{meier2021language}, ESM-2 \cite{lin2022language}, and ProtTrans \cite{elnaggar2021prottrans} learn residue representations directly from amino acid sequences through masked language modeling, demonstrating strong generalization across diverse functional prediction tasks. Structure-based models utilize explicit 3D geometric information, typically employing graph neural networks or geometric encoders such as MIF \cite{yang2023masked} and PST \cite{chen2024endowing}, to improve structure-dependent site prediction. However, both approaches are inherently limited by relying on information from a single modality.

\paragraph{Multi-modal models} address this limitation by incorporating auxiliary information from protein textual descriptions or complementary annotations. Recent approaches, including MMSite \cite{ouyang2024mmsite}, UniSite \cite{fan2025unisite}, and ProtST \cite{xu2023protst}, integrate protein–text pairs or fuse sequence and structure encoders to improve active site identification performance. Although these approaches align sequence and text features through cross-attention or end-to-end multimodal fusion strategies, most existing methods employ relatively shallow integration schemes that lack residue-level adaptive fusion and explicit reliability estimation. Consequently, they may not fully exploit the complementary strengths of each modality.

\begin{figure*}[t]
    \centering
    \includegraphics[clip,scale=0.235]{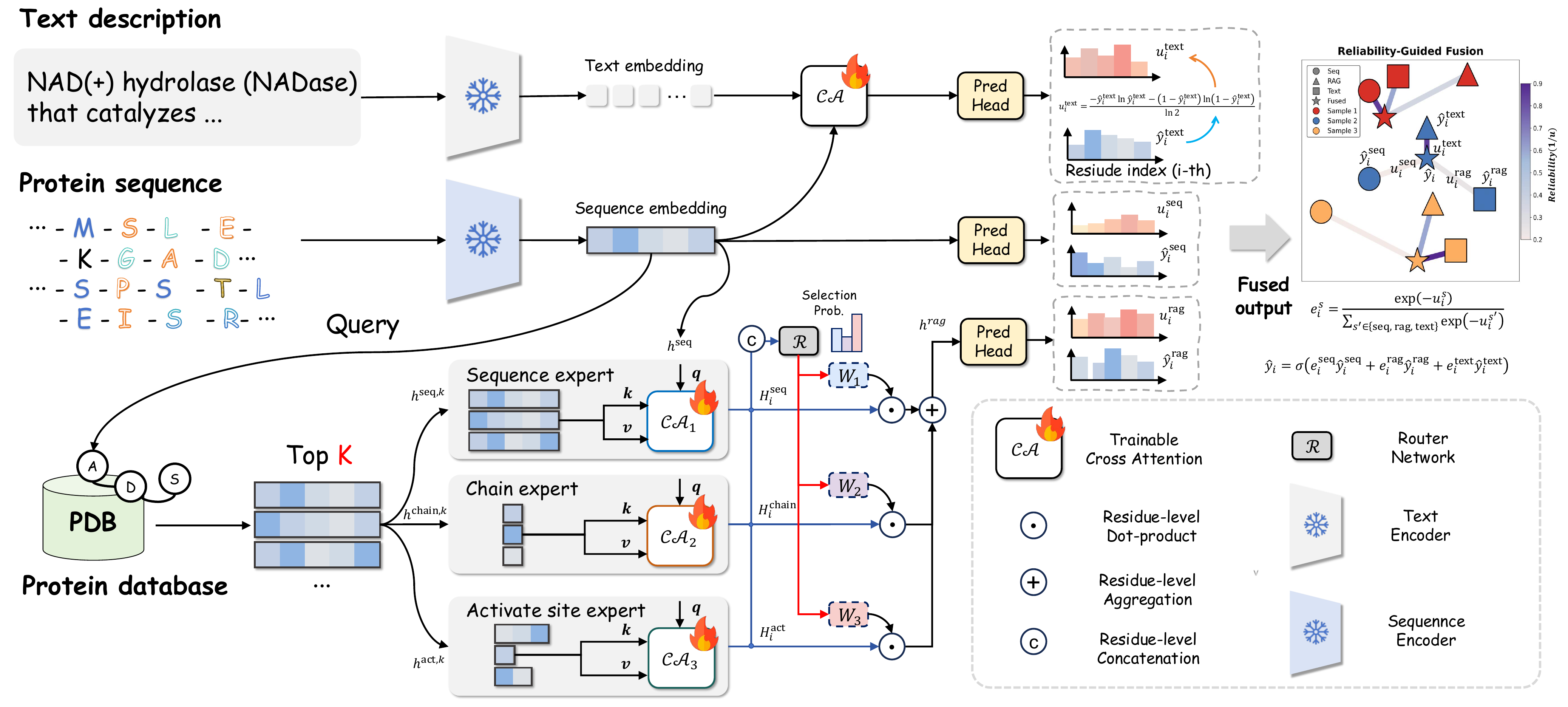}
    \vspace{-2mm}
    \caption{The framework of the proposed MERA.}
    \label{fig:framework}
    \vspace{-7mm}
\end{figure*}

\subsection{Retrieval-Augmented Generation}
Recent works have explored retrieval-augmented and multi-modal strategies to address the challenges of limited supervision and representation sparsity in protein modeling. MSM-Mut \cite{guo2024enhancing} enriches residue features by retrieving local structural motifs; RAPM \cite{wu2025rethinking} combines sequence and embedding-based retrieval for protein annotation; and ProTrek \cite{su2024protrek} enables cross-modal retrieval. These approaches demonstrate that incorporating external or multi-modal information can effectively alleviate the under-determined nature of residue representations. However, existing retrieval-augmented and multi-modal models typically employ static residue representations within rigid, task-specific pipelines. This design limitation restricts their ability to capture the localized, diverse, and context-dependent characteristics of active-site distributions, making it difficult to adaptively incorporate relevant information from multiple biological perspectives.

%% file: section/3-method.tex
\section{Methodology}

\subsection{Preliminaries}

A protein sequence is represented as $\mathcal{S} = [s_1, \dots, s_n]$ and its textual description as $\mathcal{T} = [t_1, \dots, t_m]$, where $s_i$ and $t_j$ denote amino acids and text tokens, respectively. Multi-modal active-site identification is computed as $\hat{\mathbf{y}} = f_\theta(\mathcal{S}, \mathcal{T})$, where $\hat{\mathbf{y}} = [\hat{y}_1, \dots, \hat{y}_n]$ is the output vector and each component $\hat{y}_i$ represents the likelihood that residue $s_i$ is an active site. The function $f_\theta : \mathbb{R}^{n\times 1280} \times \mathbb{R}^{m\times 768} \to \{0, 1\}^n$ is learned to perform this mapping.

\paragraph{Protein Vector Database.}
To enable retrieval-augmented modeling, we construct a protein vector database
\begin{equation*}
    \mathcal{D} = \bigl \{ (\mathbf{h}^{\text{chain},j}, \mathbf{h}^{\text{seq},j}, \mathbf{h}^{\text{act},j}) \bigr\}_{j=1}^{M},
\end{equation*}
where $M$ denotes the total number of training proteins. For each protein $j$:
\begin{itemize}
  \item $\mathbf{h}^{\text{chain},j} \in \mathbb{R}^{n_{chain} \times 1280}$ is the chain-level embedding computed as the mean of $\mathbf{h}^{\text{seq},j}$ and serves as the searchable key for database, where $n_{chain} =1$.
  \item $\mathbf{h}^{\text{seq},j} \in \mathbb{R}^{n_{seq} \times 1280}$ is the residue-level embedding with ${n_{seq}}$ residues.
  \item $\mathbf{h}^{\text{act},j} \in \mathbb{R}^{n_{act}\times 1280}$ contains $n_{act}$ residue embeddings indicated by the ground-truth active-site mask, with $1 \le n_{act} \le n_{seq}$.
\end{itemize}

\paragraph{Online Retrieval.}
At inference, given a query protein $\mathcal{S}$ with chain embedding $\mathbf{h}^{\text{chain}}$, we retrieve the top-$K$ most similar neighbors from the database
\begin{equation*}
    \{ \mathbf{h}^{e,k} \}_{k=1}^{K} = \underset{\substack{j \in \mathcal{D}, j \neq \text{query}}} {\operatorname{arg\,max}^{K}} \frac{\mathbf{h}^{\text{chain}} \cdot \mathbf{h}^{\text{chain},j}} {\lVert \mathbf{h}^{\text{chain}} \rVert \lVert \mathbf{h}^{\text{chain},j} \rVert},
\end{equation*}
where cosine similarity is utilized to measure the relatedness between query and database proteins. Each retrieved neighbor is represented as $\mathbf{h}^{e,k} = ( \mathbf{h}^{\text{chain},k}, \mathbf{h}^{\text{seq},k}, \mathbf{h}^{\text{act},k} )$.

\subsection{Framework Overview}

The overall framework is illustrated in Figure~\ref{fig:framework}. For a protein sequence $\mathcal{S}=[s_1, \dots, s_n]$, we use ESM-1b to obtain sequence-level embeddings $\mathbf{h}^{\text{seq}} \in \mathbb{R}^{n \times 1280}$ and compute a sequence key $\mathbf{h}^{\text{chain}} = \frac{1}{n} \sum_{i=1}^{n} \mathbf{h}^{\text{seq},i}$. The corresponding textual description $\mathcal{T}$ is first encoded by BioMedBERT into a text embedding $\tilde{\mathbf{h}}^{\text{text}} \in \mathbb{R}^{m \times 768}$. We then apply a cross-attention module to generate text-guidance embeddings for each residue, producing a text-enhanced residue representation $\mathbf{h}^{\text{text}} \in \mathbb{R}^{n \times 1280}$ aligned in length with sequence embedding $\mathbf{h}^{\text{seq}}$ to facilitate subsequent fusion.
Next, we query the database with $\mathbf{h}^{\text{chain}}$ and retrieve the top-$K$ neighbors $\{ \mathbf{h}^{e,k} \}_{k=1}^{K}$. These neighbors and $\mathbf{h}^{\text{seq}}$ are fed into a Multi-expert RAG (MeRAG) module to produce an enhanced representation $\mathbf{h}^{\text{rag}} \in \mathbb{R}^{n \times 1280}$ that aggregates information from three complementary perspectives: sequence, chain, and active-site experts.
The Reliability-aware Multimodal Fusion (RMF) module dynamically evaluates the reliability of three modalities for each residue: the original sequence modality $\mathbf{h}^{\text{seq}}$, RAG-enhanced $\mathbf{h}^{\text{rag}}$, and textual-enhanced $\mathbf{h}^{\text{text}}$. This enables the model to assign higher importance to more confident modalities at each position, resulting in a fused residue-level representation $\mathbf{h}^{\text{fused}} \in \mathbb{R}^{n \times 1280}$ that effectively balances information from all perspectives according to their estimated trustworthiness. This representation is finally mapped to the active-site probability vector $\hat{\mathbf{y}} \in [0,1]^{n}$.

\subsection{Multi-expert RAG (MeRAG)}  
\label{sec:multi_expert}

This section describes our Multi-expert RAG (MeRAG) module that leverages multiple expert perspectives to enhance residue-level representations through retrieval-augmented generation. Given a query protein sequence $\mathcal S$ with $n$ residues, we extract sequence embeddings $\mathbf h_{\text{seq}}$ and the chain-level key $\mathbf h_{\text{chain}}$ to retrieve the top-$K$ neighbors $\{ \mathbf h^{e,k} \}_{k=1}^{K}$, where each neighbor $k$ consists of $\mathbf h^{\text{seq},k}$, $\mathbf h^{\text{chain},k}$ and $\mathbf h^{\text{act},k}$.

\paragraph{Per-expert neighbor summarization.}  
Three experts \emph{seq}, \emph{chain}, and \emph{act} operate in parallel on every residue $i$ of the query across the neighbors:  
\begin{equation*}
    \mathbf{H}^{e} = \operatorname{Expert}_{e} \bigl( \mathbf{h}^{\text{seq}},\{\mathbf{h}^{e,k}\}_{k=1}^{K} \bigr),\quad e \in \{ \text{seq}, \text{chain}, \text{act} \},
\end{equation*} 
yielding expert outputs {\small$\mathbf H^{e} \in \mathbb{R}^{n\times 1280}$} for calculating the retrieval-augmented enhancements from three perspectives.
Rather than applying cross-attention directly between the query and all neighbor residues, we adopt a hierarchical (intra/inter) aggregation scheme that first denoises and summarizes local context within each neighbor before global integration. This design, inspired by recent advances in protein modeling and graph learning \cite{jumper2021highly,fout2017protein,borgeaud2022improving}, reduces computational cost and overfitting while yielding more robust and interpretable retrieval-augmented representations.

\paragraph{Intra-neighbor aggregation.}  
For each query residue $i$ and retrieved neighbor $k = 1, \dots, K$, we aggregate the $n_k$ residue embeddings of that neighbor into a single vector $\mathbf{z}_i^{e,k} \in \mathbb{R}^{1 \times 1280}$. Specifically, we weight every residue $j$ within neighbor $k$ by its similarity to query residue $i$:  
\begin{equation*}
    \mathbf{z}_i^{e,k} = \sum_{j=1}^{n_k} \beta_{ij}^{e,k} \mathbf{h}_j^{e,k}, \quad \beta_{ij}^{e,k} = \frac{ \exp \bigl( \mathbf{h}^{\text{seq},i} \cdot \mathbf{h}_j^{e,k}/\tau \bigr) }{\sum_{j'=1}^{n_k} \exp \bigl( \mathbf{h}^{\text{seq},i} \cdot \mathbf{h}_{j'}^{e,k}/\tau \bigr) },
\end{equation*}
where $\tau = 0.1$ is a temperature parameter that sharpens the attention weights $\beta_{ij}^{e,k}$, $j$ indexes residues within neighbor $k$, and $n_k$ is the length of neighbor $k$.

\paragraph{Inter-neighbor fusion.}  
After obtaining $K$ neighbor-level summaries $\{\mathbf z_i^{e,k}\}_{k=1}^{K}$, we fuse them with the query residue representation into a single expert output $\mathbf{H}_i^{e} \in \mathbb{R}^{1 \times 1280}$. Specifically, we treat the query embedding as the $0$-th candidate and compute a weighted sum over all $K+1$ candidates:
\begin{equation*}
    \mathbf{H}_i^{e} = \sum_{k=0}^{K}\gamma_{ik}^{e} \mathbf{z}_i^{e,k}, \quad \gamma_{ik}^{e} = \frac{ \exp \bigl( \mathbf{h}^{\text{seq}}_i \cdot \mathbf{z}_i^{e,k} \bigr) }{\sum_{k'=0}^{K} \exp \bigl( \mathbf{h}^{\text{seq}}_i \cdot \mathbf{z}_i^{e,k'} \bigr) },
\end{equation*}  
where $\gamma_{ik}^{e}$ quantifies the contribution of neighbor $k$ to the final expert output $\mathbf{H}_i^{e}$ and index $k=0$ corresponds to the query residue itself, with $\mathbf{z}_i^{e,0} = \mathbf{h}^{\text{seq}}_i$.

\paragraph{Mixture-of-Experts gating.}
The three expert outputs are fed into an MoE layer to obtain the residue-level RAG-enhanced representation $\mathbf{h}^{\text{rag}} \in \mathbb{R}^{n \times 1280}$ via a residue-wise soft gate: 
\begin{equation*}
    \begin{aligned}
        \mathbf{h}^{\text{rag}}_i &= \operatorname{MoE} \bigl( \mathbf{H}_i^{\text{seq}},\mathbf{H}_i^{\text{chain}},\mathbf{H}_i^{\text{act}} \bigr) = \sum_{e}g_i^{e} \mathbf{H}_i^{e}, \\
        \mathbf{g}_i &= \operatorname{Softmax} \Bigl( \operatorname{MLP}_\theta \bigl( \left[ \mathbf{H}_i^{\text{seq}}; \mathbf{H}_i^{\text{chain}}; \mathbf{H}_i^{\text{act}} \right] \bigr) \Bigr) \in \mathbb{R}^{3 \times 1280},
\end{aligned}
\end{equation*}
where $[\cdot; \cdot; \cdot]$ denotes concatenation and $\operatorname{MLP}_\theta$ is a two-layer MLP that outputs expert selection probabilities.

This residue-level gating mechanism allows the model to adaptively integrate three expert perspectives at each residue position, enabling finer-grained fusion than applying a single global MoE to the entire sequence. This fine-grained adaptation is particularly important in capturing heterogeneous contributions of different experts across the protein, which is essential for modeling the diverse local contexts present in protein sequences. This approach is consistent with previous findings in token-wise adaptive gating in large-scale neural networks \cite{shazeer2017outrageously,riquelme2021scaling}.

\subsection{Reliability-aware Multimodal Fusion (RMF)}

We now present our Reliability-aware Multimodal Fusion (RMF) module that employs evidence-theoretic methods to quantify modality reliability and achieve principled multimodal integration. Our framework integrates three complementary modalities: the sequence representation $\mathbf{h}^{\text{seq}} \in \mathbb{R}^{n \times 1280}$, the RAG-enhanced representation $\mathbf{h}^{\text{rag}} \in \mathbb{R}^{n \times 1280}$, and the text-guided representation $\mathbf{h}^{\text{text}} \in \mathbb{R}^{n \times 1280}$, where $n$ denotes the number of residues. Inspired by Dempster–Shafer evidence theory \cite{amini2020deep,huang2025deep,deregnaucourt2025conflict}, we model each modality's reliability through belief mass functions and apply learnable discounting coefficients to enable principled fusion at each residue position.

\paragraph{Modality-specific prediction head.}
To enable reliability-aware fusion based on evidence theory, we deploy three parallel prediction heads, each specialized for a specific modality. For each modality $s \in \{\text{seq}, \text{rag}, \text{text}\}$, we define a modality-specific prediction head $\mathrm{Pred}_s$ that maps the residue representation $\mathbf{h}^{s}_i$ to a prediction logit $\hat{y}^s_i \in [0,1]$:
\begin{equation*}
    \hat{y}^s_i = \mathrm{Pred}_s(\mathbf{h}^{s}_i), \quad s \in \{\text{seq}, \text{rag}, \text{text}\},\ i = 1, \ldots, n.
\end{equation*}
Each $\mathrm{Pred}_s$ is implemented as a two-layer MLP with independent parameterization across modalities to capture modality-specific predictive patterns.

\paragraph{Modality Reliability Estimation.}
Following Dempster–Shafer evidence theory, we model each modality's prediction as a belief mass function and quantify its reliability through learnable discounting. For each residue $i$, the evidence mass is computed as:
\begin{equation*}
    m^s_i = \frac{\exp(\hat{y}^s_i)}{\sum_{s'} \exp(\hat{y}^{s'}_i)}, \quad s \in \{\text{seq}, \text{rag}, \text{text}\},\ i = 1,\ldots,n,
\end{equation*}
where $m^s_i$ the relative evidence strength of modality $s$ for residue $i$.

To assess the modality trustworthiness beyond simple prediction confidence, we compute a credibility (discounting) coefficient following~\cite{han2024fmfn}:
\begin{equation*}
    c^s_i = \frac{1}{2} \left( m^s_i + 1 - \max_{s' \ne s} m^{s'}_i \right),
\end{equation*}
where $c^s_i \in [0,1]$ serves as a learnable discounting factor that encourages high reliability only when modality $s$ not only achieves strong evidence but also distinguishes itself from competing modalities. This design prevents unreliable modalities from dominating the fusion, addressing the core limitation of naive logit-based weighting.

\paragraph{Reliability Quantification.}
To convert credibility coefficients into fusion weights, we measure the reliability indicator $u^s_i \in [0,1]$ for residue $i$ in modality $s$ as the normalized binary entropy of its credibility coefficient:
\begin{equation*}
    u^s_i = \frac{ -c^s_i \ln c^s_i - (1-c^s_i) \ln (1-c^s_i) }{ \ln 2 },
\end{equation*}
where lower $u^s_i$ values indicate higher reliability for modality $s$ at residue $i$.

\paragraph{Reliability-guided Adaptive Fusion.}
At each residue $i$, we perform adaptive fusion using normalized weights derived from their reliability indicators:
\begin{equation*}
    e^{s}_i = \frac{ \exp(-u^s_i) }{ \sum_{s' \in \{\text{seq}, \text{rag}, \text{text}\}} \exp(-u^{s'}_i) },
\end{equation*}
where $e^s_i$ is the evidence-based fusion weight for modality $s$ at residue $i$. 
The final prediction is computed as a reliability-weighted combination of the logits from all modalities:
\begin{equation*}
    \hat{y}_i = \sigma \left( e^{\text{seq}}_i \hat{y}^{\text{seq}}_i + e^{\text{rag}}_i \hat{y}^{\text{rag}}_i + e^{\text{text}}_i \hat{y}^{\text{text}}_i \right), \quad i=1, \ldots, n,
\end{equation*}
where $\hat{y}^s_i$ denotes the logit from modality $s$, $e^s_i$ denotes the reliability-based weight, and $\sigma$ represents the element-wise sigmoid function.

\subsection{Training Objective}
The model is trained using a binary cross-entropy loss applied to the final fused prediction $\hat{y}_i$:
\begin{equation*}
    \mathcal{L}_{\text{bce}} = -\frac{1}{n}\sum_{i=1}^{n} \left[ y_i\log\hat{y}_i + (1-y_i)\log(1-\hat{y}_i) \right].
\end{equation*}
A reliability regularization term encourages each modality-specific prediction to align with the ground truth:
\begin{equation*}
    \mathcal{L}_{\text{reliability}} = \sum_{s} \sum_{i=1}^{n} |\hat{y}^s_i - y_i|^2.
\end{equation*}
The total training objective is
\begin{equation*}
    \mathcal{L} = \mathcal{L}_{\text{bce}} + \mathcal{L}_{\text{reliability}}.
\end{equation*}

%% file: section/4-experiments.tex
\section{Experiments}
\label{sec:experiment}

\subsection{Experimental Settings}

\noindent \textbf{Datasets.}
The ProTAD dataset introduced by MMSite \cite{ouyang2024mmsite} provides protein–text pairs with manually curated UniProt annotations. While these human-written descriptions offer high information density, their scarcity for newly discovered proteins limits ProTAD's applicability in real-world scenarios, potentially overestimating model performance. To bridge this gap, we introduce ProTAD-Gen, a challenging extension of ProTAD where all textual descriptions are automatically generated by ESM2Text. Crucially, the residue-level active-site labels remain identical to ProTAD, enabling fair performance comparison. By replacing human curation with automated text generation, ProTAD-Gen provides a more realistic benchmark for evaluating model robustness. 
To prevent data leakage, we preprocess ProTAD-Gen by cleaning and filtering the raw data, followed by MMseqs2 clustering at 10\% sequence identity threshold \cite{steinegger2017mmseqs2}. To further validate the generalization capability of our framework, we also evaluate on the TS125 dataset, which is annotated with peptide-binding residues \cite{taherzadeh2016sequence}. Compared to single-sequence identification in ProTAD-Gen, binding site identification on TS125 protein–peptide complexes poses a much greater challenge. We generate textual descriptions for all TS125 samples using ESM2Text. The preprocessing steps for both datasets follow PepCA \cite{huang2024pepca}, including removal of redundant sequences using BLAST \cite{boratyn2013blast} and MMseqs2 (with a 30\% sequence identity threshold). For both datasets, we use an 8:1:1 train/validation/test split, resulting in 5,569/697/697 samples for ProTAD (6,963 proteins), and 1,040/75/75 samples for TS125 (1,190 proteins).

\paragraph{Baselines.}
We evaluate MERA against state-of-the-art methods on two tasks:
\begin{itemize}
    \item \textbf{Active site identification (ProTAD-Gen):} we compare against \textit{sequence-only models}, including ESM-1b~\cite{rives2021biological}, ESM-1v~\cite{meier2021language}, ESM-2~\cite{lin2022language}, ProtElectra~\cite{elnaggar2021prottrans}, PETA~\cite{tan2024peta}, TAPE~\cite{rao2019evaluating}, and S-PLM~\cite{wang2025s}; \textit{structure-aware models} PST~\cite{chen2024endowing} and MIF~\cite{yang2023masked}; and \textit{multi-modal models} MMSite~\cite{ouyang2024mmsite}, ProtST~\cite{xu2023protst}, and UniSite~\cite{fan2025unisite}.
    \item \textbf{Peptide-binding site prediction (TS125):} We evaluate against \textit{sequence-based deep models} IIDL-PepPI~\cite{chen2025protein}, PepCA~\cite{huang2024pepca}, PepBCL~\cite{wang2022predicting}, PepCNN~\cite{chandra2023pepcnn}, and PepNN~\cite{abdin2022pepnn};  \textit{protein language models} ESM-2~\cite{lin2022language} and TAPE~\cite{rao2019evaluating}; \textit{structure-aware methods:} PepBind~\cite{zhao2018improving} and PepSite~\cite{petsalaki2009accurate}.
\end{itemize}
Results of all baselines are reproduced using publicly available code implementations with default hyperparameters to ensure fair evaluation.

\paragraph{Implementation.}
Protein sequences are encoded using ESM-1b (1280-dimensional embeddings), while textual descriptions are processed by BioMedBERT (768-dimensional embeddings). For inference, the original MMSite framework requires UniProt IDs to generate protein descriptions via Prot2Text. To enable broader applicability, particularly for novel proteins lacking UniProt annotations, we replace this with sequence-to-text models like ESM2Text, which directly generates functional annotations solely from raw sequence data.
For retrieval, we select the top-$3$ nearest neighbors within each cluster, achieving an effective balance between computational efficiency and retrieval performance, as validated by our ablation studies. All experiments are conducted on two 40 GB NVIDIA A100 GPUs. Models are trained for 100 epochs, requiring approximately three hours for each task. For the TS125 dataset, we extend our framework with an additional peptide expert. We use the Adam optimizer with a learning rate of $1 \times 10^{-3}$. Our model contains approximately 300M parameters.

\paragraph{Evaluation.}
The model checkpoint achieving the highest AUPRC on the validation set is selected for final evaluation.  During evaluation, we report F\textsubscript{max}, AUPRC, AUROC, MCC, and Hits@k (where $k=1,5,10$) metrics on the test set. F\textsubscript{max} denotes the maximum F1-score across varying probability thresholds, indicating the best trade-off between precision and recall. AUPRC summarizes the precision-recall relationship and is particularly informative for imbalanced datasets. AUROC, used for TS125, measures the model's ability to distinguish between binding and non-binding residues. MCC provides a balanced assessment of prediction quality by considering all four confusion matrix categories. Hits@k reflects the proportion of true active-site or binding-site residues ranked among the top $k$ predictions for each protein, evaluating the model's ability to prioritize true sites at the top of its output.

\input{tables/exp1.tex}

\subsection{Main Results}

\begin{figure}[!t]
    \centering
    \includegraphics[clip,scale=0.21]{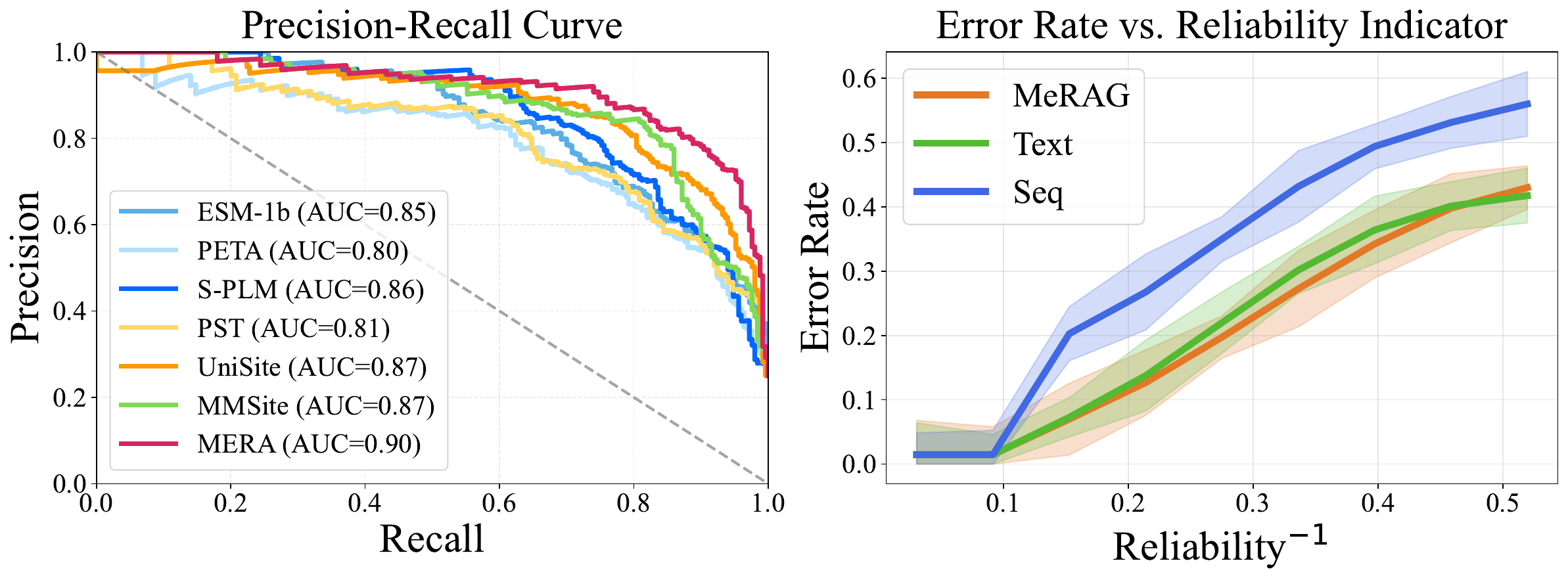}
    \caption{Comparison on ProTAD-Gen. \textbf{Left:} Precision recall curves. \textbf{Right:} Error rate vs. reliability indicator under high-confidence predictions ({\small$\hat{y} > 0.8$ or $\hat{y} < 0.2$}).}
    \label{fig:roc}
\end{figure}

\textbf{Key Observation 1: Superior and robust active-site identification across benchmarks.}
MERA consistently achieves the best results on both ProTAD-Gen and TS125 datasets. On ProTAD-Gen, MERA obtains an AUPRC of 0.90 and F$_\mathrm{max}$ of 0.88, representing 3\% and 7\% improvements over \textbf{MMSite}, respectively. On the more challenging TS125 dataset, MERA achieves the highest AUROC of 0.85, demonstrating strong cross-task generalization capability. This superior performance is attributed to the use of retrieval-augmented residue representations and adaptive multimodal fusion based on reliability, which together yield more discriminative identification of protein active sites. Visualization results (Figure~\ref{fig:roc} (a)) further confirm MERA's consistent advantage in precision-recall curves across all thresholds.
\textbf{Key Observation 2: Strong real-world utility and biological relevance.}
MERA demonstrates substantial improvements in ranking metrics, achieving 0.98 Hits@10 score and increasing Hits@1 by 3\% on ProTAD-Gen and 6\% on TS125 compared to the strongest baselines. This enables rapid and confident prioritization of candidate sites for experimental validation, substantially reducing the time and cost requirements for wet-lab studies.
\textbf{Key Observation 3: Reliability estimation enables trustworthy multi-modal fusion.}
Our reliability estimation provides a robust measure of prediction trustworthiness across modalities. Figure~\ref{fig:roc} (b) illustrates a consistent monotonic relationship between reliability indicators and error rates for high-confidence predictions {\small($\hat{y} > 0.8$ or $\hat{y} < 0.2$)} across three modalities. Lower reliability consistently corresponds to higher error rates, validating that our reliability quantification offers a principled criterion for assessing prediction quality. Unlike raw predicted probabilities that may poorly reflect true model confidence, our reliability-aware approach provides calibrated measures that enable trustworthy multimodal fusion decisions.

\input{tables/exp2.tex}

\subsection{Case study} 

\begin{figure}[!t]
    \centering
    \includegraphics[clip,scale=0.65]{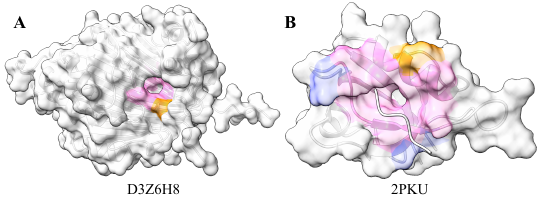}
    \caption{Colors on the surface (residues) indicate correctly predicted sites (pink), unpredicted sites (yellow), and incorrectly predicted sites (blue).}
    \label{fig:case}
\end{figure}

We select two representative cases from different test sets. For S-adenosylmethionine decarboxylase proenzyme 2 (UniProt: D3Z6H8), MERA successfully identified most active sites but missed one position at the edge of the binding region. This oversight is visualized in the Figure~\ref{fig:case} A using AlphaFold 3, highlighting the structural context of the missed site. Despite this minor error, the overall prediction aligns well with the known functional residues. For the PICK1 PDZ domain (PDB: 1PKU), which involves peptide binding. MERA correctly predicted the majority of the active sites in the primary binding region, as shown in Figure~\ref{fig:case}B. This demonstrates its effectiveness in capturing key interaction sites. However, some peripheral positions were either unpredicted or incorrectly identified, reflecting potential limitations in capturing less critical or structurally ambiguous regions. 
These cases provide valuable insights into the strengths and limitations of MERA, particularly in handling edge cases and distinguishing between essential and secondary binding sites. Overall, these examples underscore the robustness of our approach while identifying areas for future improvement.

\input{tables/ablation}

\subsection{Ablation Study}

\textbf{Contribution of Each Module.} 
Table~\ref{tab:ablation} presents the ablation results for different modules on the ProTAD-Gen dataset.
\textbf{Key Observation 1: Multimodal fusion is essential but requires principled integration.} Removing the RMF module entirely leads to the most severe performance degradation (AUPRC drops from 0.90 to 0.83), confirming that naive single-modality approaches are fundamentally inadequate. When modalities are integrated without proper reliability assessment, performance can degrade significantly, highlighting the importance of our reliability-aware fusion strategy.
\textbf{Key Observation 2: Introducing additional modalities substantially enhances sequence-only predictions.} Both text and RAG modalities provide substantial improvements over the sequence-only baseline, demonstrating the effectiveness of incorporating auxiliary information sources.
\textbf{Key Observation 3: MeRAG fusion enables effective expert integration.} Replacing our residue-level MeRAG fusion with direct expert combination leads to notable performance degradation, indicating that naive fusion of RAG experts introduces noise and fails to capture position-specific contributions. Our MeRAG mechanism is crucial for effectively utilizing retrieval-augmented information through adaptive expert weighting at each residue position.
\textbf{Key Observation 4: Each expert provides complementary retrieval perspectives.} Removing any single expert consistently degrades performance, confirming the complementarity of multi-expert retrieval approach.

\begin{figure}[t]
    \centering
    \includegraphics[clip,scale=0.20]{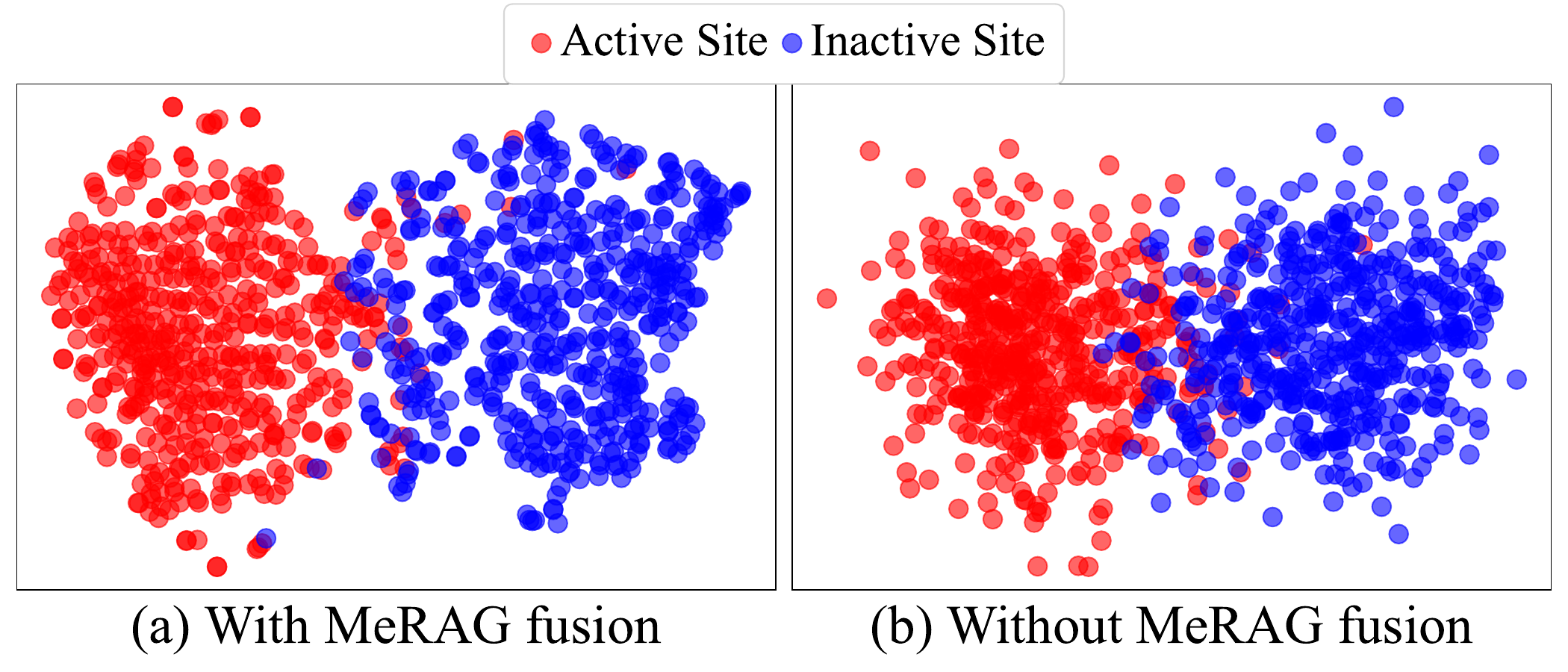}
    \vspace{-2mm}
    \caption{Visualization of inactive (blue) and active (red) site embeddings w/ and w/o MeRAG fusion on ProTAD-Gen.}
    \label{fig:embeddings}
\end{figure}

\paragraph{Visualization of Embedding Discriminability.}
To further visualize the distinguishability of active and inactive sites with and without the MeRAG module, we randomly sampled 500 active-site residues and 500 inactive-site residues from the test set and projected them into a 2D space for comparison. As shown in Figure~\ref{fig:embeddings}, the full MeRAG framework produces much clearer separation and clustering between active (red) and inactive (blue) sites, whereas removing the MeRAG module leads to increased overlap between the two classes. This demonstrates the advantage of our approach in learning more discriminative residue-level representations.

\input{tables/four}

\paragraph{Evaluating Flexibility of MERA.}
We conduct an ablation study on the TS125 dataset (Table~\ref{tab:ts125_ablation}) to demonstrate the flexibility and generalization of our approach in complex peptide–protein binding scenarios. The results show that incorporating a specialized peptide expert (four experts total) further enhances the performance compared to the three-expert variant. This improvement underscores the adaptability of the MERA framework, as new experts can be flexibly integrated to handle diverse biological tasks.
Importantly, even with only three experts, our model still outperforms the strongest baseline (IIDL-PepPI) across all evaluation metrics, further highlighting the robustness and strong generalization ability of our framework.

%% file: tables/exp1.tex
\begin{table}[t]
\small
\centering
\vspace{-2mm}
\resizebox{\columnwidth}{!}{%
  \renewcommand{\arraystretch}{0.8}
  \setlength{\tabcolsep}{3pt}
  \begin{tabular}{lccccccc}
    \toprule
    \textbf{Method} & F\textsubscript{max} & AUPRC & MCC & Hits@1 & Hits@5 & Hits@10 \\
    \midrule
    ESM-1b        & 0.71 & 0.85 & 0.71 & 0.76 & 0.88 & 0.90 \\
    ESM-1v        & 0.63 & 0.80 & 0.64 & 0.68 & 0.81 & 0.84 \\
    ESM-2-650M    & 0.65 & 0.82 & 0.66 & 0.70 & 0.83 & 0.86 \\
    ProtElectra   & 0.56 & 0.76 & 0.57 & 0.59 & 0.74 & 0.79 \\
    PETA          & 0.65 & 0.80 & 0.66 & 0.70 & 0.84 & 0.87 \\
    S-PLM         & 0.73 & 0.86 & 0.73 & 0.78 & 0.89 & 0.91 \\
    TAPE          & 0.36 & 0.54 & 0.36 & 0.38 & 0.56 & 0.61 \\
    MIF           & 0.14 & 0.35 & 0.14 & 0.16 & 0.29 & 0.34 \\
    ProtST        & 0.46 & 0.70 & 0.47 & 0.51 & 0.67 & 0.72 \\
    PST           & 0.66 & 0.81 & 0.66 & 0.71 & 0.84 & 0.87 \\
    UniSite       & \underline{0.82} & \underline{0.87} & 0.81 & 0.86 & \underline{0.94} & \underline{0.95} \\
    MMSite        & 0.81 & \underline{0.87} & \underline{0.83} & \underline{0.88} & \underline{0.95} & \underline{0.95} \\
    \midrule
    \textbf{MERA} & \textbf{0.88} & \textbf{0.90} & \textbf{0.88} & \textbf{0.91} & \textbf{0.97} & \textbf{0.98} \\
    \bottomrule
  \end{tabular}%
}
\label{table1}
\caption{Results on the ProTAD-Gen dataset.}
\end{table}

%% file: tables/exp2.tex
\begin{table}[t]
\small
\centering
\vspace{-2mm}
\resizebox{\columnwidth}{!}{%
  \renewcommand{\arraystretch}{0.8}
  \setlength{\tabcolsep}{3pt}
  \begin{tabular}{lcccccc}
    \toprule
    \textbf{Method} & F\textsubscript{max} & AUROC & MCC & Hits@1 & Hits@5 & Hits@10 \\
    \midrule
    Pepsite    & 0.30 & 0.61 & 0.30 & 0.35 & 0.62 & 0.70 \\
    PepCNN     & 0.33 & 0.68 & 0.33 & 0.40 & 0.67 & 0.75 \\
    PepBind    & 0.34 & 0.79 & 0.34 & 0.42 & 0.72 & 0.78 \\
    PepNN      & 0.31 & 0.73 & 0.31 & 0.39 & 0.68 & 0.76 \\
    TAPE       & 0.30 & 0.75 & 0.30 & 0.41 & 0.70 & 0.77 \\
    ESM2       & 0.32 & 0.77 & 0.32 & 0.43 & 0.72 & 0.80 \\
    PepBCL     & 0.33 & 0.79 & 0.33 & 0.44 & 0.73 & 0.81 \\
    PepCA      & 0.33 & 0.83 & 0.34 & 0.43 & 0.76 & 0.84 \\
    IIDL-PepPI & \underline{0.35} & \underline{0.84} & \underline{0.35} & \underline{0.46} & \underline{0.77} & \underline{0.85} \\
    \midrule
    \textbf{MERA} & \textbf{0.40} & \textbf{0.85} & \textbf{0.41} & \textbf{0.52} & \textbf{0.79} & \textbf{0.86} \\
    \bottomrule
  \end{tabular}%
}
\caption{Results on the TS125 dataset.}
\label{table2}
\vspace{-3mm}
\end{table}

%% file: tables/ablation.tex
\begin{table}[t]
\small
\centering
\renewcommand{\arraystretch}{0.80}
\setlength{\tabcolsep}{3pt}
\vspace{-2mm}
\resizebox{\columnwidth}{!}{%
\begin{tabular}{lccccc}
\toprule
Method                & F$_\mathrm{max}$ & AUPRC & MCC   & Hits@1 & Hits@10 \\
\midrule
MERA               & \textbf{0.88}   & \textbf{0.90}  & \textbf{0.88}  & \textbf{0.91} & \textbf{0.98} \\
w/o RMF      & 0.70   & 0.83  & 0.70  & 0.75 & 0.88 \\
w/o Text modality      & 0.83   & 0.86  & 0.83  & 0.87 & 0.94 \\
w/o RAG modality      & 0.79   & 0.86  & 0.80  & 0.86 & 0.92 \\
w Seq Modality Only   & 0.71   & 0.85  & 0.71  & 0.76 & 0.90 \\
w/o MeRAG        & 0.76   & 0.85  & 0.76  & 0.84 & 0.90 \\
w/o Sequence Expert        & 0.83   & 0.87  & 0.84  & 0.89 & 0.95 \\
w/o Chain Expert        & 0.84   & 0.86  & 0.83  & 0.87 & 0.93 \\
w/o Active-site Expert        & 0.85   & 0.87  & 0.84  & 0.90 & 0.95 \\
\bottomrule
\end{tabular}
}
\caption{Ablation study on the ProTAD-Gen dataset.}
\label{tab:ablation}
\end{table}

%% file: tables/four.tex
\begin{table}[t]
\small
\centering
\vspace{-2mm}
\resizebox{\columnwidth}{!}{%
  \renewcommand{\arraystretch}{0.9}
  \setlength{\tabcolsep}{3pt}
  \begin{tabular}{lcccc}
    \toprule
    Method                  & F$_\mathrm{max}$ & AUROC & MCC   & Hits@10 \\
    \midrule
    IIDL-PepPI              & 0.35 & 0.84 & 0.35 & 0.85 \\
    MERA (4 experts)        & \textbf{0.40}   & \textbf{0.85}  & \textbf{0.41}  & \textbf{0.86} \\
    w/o Peptide Expert (3 experts) & 0.37 & 0.84 & 0.38 & 0.83 \\
    \bottomrule
  \end{tabular}%
}
\caption{Ablation study of the number of expert modules on the TS125 dataset.}
\label{tab:ts125_ablation}
\vspace{-3mm}
\end{table}

%% file: section/5-conclusion.tex
\section{Conclusion}
\label{sec:conclusion}
We introduced MERA, the first retrieval-augmented framework for protein active-site prediction that employs residue-level mixture-of-experts. By coupling hierarchical multi-expert retrieval with reliability-aware multimodal fusion, MERA addresses the dual challenges of single-instance prediction vulnerability and inadequate modality reliability estimation that plague current methods. The framework not only achieves the best performance on benchmarks, but also demonstrates unprecedented flexibility. Future work will extend MERA to incorporate additional modalities, notably 3D structural information \cite{kim2025easy}, by introducing dedicated structure experts within the MeRAG module.